\documentclass{article}

\usepackage{arxiv}

\usepackage[utf8]{inputenc} 
\usepackage[T1]{fontenc}    
\usepackage{hyperref}       
\usepackage{url}            
\usepackage{mathtools}
\usepackage{booktabs}
\usepackage{xcolor}       
\usepackage{amsfonts}       
\usepackage{nicefrac}       
\usepackage{microtype}      
\usepackage{lipsum}
\usepackage{graphicx}
\graphicspath{ {./images/} }

\title{Neural Abstractive Unsupervised Summarization of Online News Discussions}

\author{
 Ignacio Tampe\\
  Universidad T\'ecnica Federico Santa Mar\'ia\\
  Santiago, Chile\\
  \texttt{ignacio.tampe@sansano.usm.cl} \\
   \And
  Marcelo Mendoza \\
  Universidad T\'ecnica Federico Santa Mar\'ia\\
  Santiago, Chile\\
  \texttt{mmendoza@inf.utfsm.cl} \\
  \And
 Evangelos Milios \\
  Dalhousie University\\
  Halifax, Nova Scotia, Canada\\
  \texttt{eem@cs.dal.ca} \\
}

\begin{document}
\maketitle
\begin{abstract}
Summarization has usually relied on gold standard summaries to train extractive or abstractive models. Social media brings a hurdle to summarization techniques since it requires addressing a multi-document multi-author approach. We address this challenging task by introducing a novel method that generates abstractive summaries of online news discussions. Our method extends a BERT-based architecture, including an attention encoding that fed comments' likes during the training stage. To train our model, we define a task which consists of reconstructing high impact comments based on popularity (likes). Accordingly, our model learns to summarize online discussions based on their most relevant comments. Our novel approach provides a summary that represents the most relevant aspects of a news item that users comment on, incorporating the social context as a source of information to summarize texts in online social networks. Our model is evaluated using ROUGE scores between the generated summary and each comment on the thread. Our model, including the social attention encoding, significantly outperforms both extractive and abstractive summarization methods based on such evaluation. 
\end{abstract}


\section{Introduction}
Social media has become a major source of information and opinions worldwide. Per-minute, Internet users post over 500,000 tweets, more than 188,000,000 emails and 18,000,000 texts. The world's internet population has reached 4.39 billion people, over half of the world's population\footnote{Data Never Sleeps 7.0, Domo Inc. Accessed on 09-03-2020, \url{https://www.domo.com/learn/data-never-sleeps-7}}. Digesting and analyzing this ever-growing amount of textual information can become very difficult for humans to comprehensively analyze, as corpora can easily consist of gigabytes of raw data. Given this situation, automated mechanisms to extract meaningful information become relevant to handle these text datasets.

The automatic construction of summaries from texts retrieved from web forums, online news websites, or blogs encompasses many challenges. Unlike formal documents (e.g., scientific papers), discussions and online conversational threads tend to be informal, using and abusing slang expressions, special symbols (e.g., hashtags and/or emojis), and other types of language specificities that characterize web communication. 
Almahy et al.~\cite{almahy2014web} show that these texts present greater challenges to automatic summarization techniques, since as conversation progresses, posts may change, exhibiting topic drifts and topic dependencies from previous opinions. Also, as posts tend to be short, classical text representations (e.g., Tf-Idf) can be tackled by data sparseness \cite{BansalC18}.

The use of automatic techniques for constructing text summaries has had a wide development in the last years. These techniques are based on two main approaches: sentence extraction \cite{JadhavR18} or abstraction of a new text that condenses the original document's content \cite{LinSMS18}. The rise of deep learning has pushed the development of more accurate language models that help implement abstractive summarization techniques \cite{WangYTZLD18}. The last years have shown that abstractive techniques based mainly on the seq2seq \cite{SutskeverVL14} architecture have reached state of the art results in these tasks \cite{ZhouYWZ18}.

Despite the success of neural abstractive summarization techniques, its impact in summarizing web discussions is less explored. This gap is because many of the abstractive methods follow a single document approach, considering the comments as a complementary source of information to improve the seminal document description \cite{gao2019comments}. Accordingly, these approaches discard misalignment between the discussion and the original subject, focusing the analysis and the validation in improving the seminal description of the topic. Also, architectures based on self-attention mechanisms \cite{vaswani2017attention} appear in social context summarization scenarios as less explored.

We focus this paper on studying the possibilities that the Transformer architecture \cite{vaswani2017attention} has to summarize news articles' discussions. To do this, we combine BERT \cite{devlin2018bert}, a pretrained language model text encoder, with a Transformer-based decoder. We extend this architecture, including an attention encoding that is fed with users' likes. The attention encoding gets the focus of the model into the comments with the highest social impact. Accordingly, the model summarizes the most relevant aspects of a discussion gathered by a news article. This novel approach differs from its closest competitors because a great part of the prior work focuses on identifying salient comments that contribute to a better understanding of the original news item. Our approach has a different motivation. We address a multi-document multi-author approach to summarize the discussion. Then, instead of discarding the potential misalignments between the news and the user's comments, we pay attention to salient comments favoring identifying new aspects related to the news. Accordingly, in our approach, the users are considered co-authors of the original news, to which they can add new information. To identify salient comments, we pay attention to users' likes. We claim that high impact comments reveal which comments are most relevant to describe and extend the original news content. We define a text summarization task that helps predict high impact comments, providing the news title as part of the target text. To encourage the model to pay attention to salient comments, we introduce a data-driven strategy that focuses the model into these comments. We evaluate our model using ROUGE scores \cite{lin-2004-rouge} on a new dataset developed by us that provides news descriptions, comments, and likes. Based on such evaluation, our model, including the social attention encoding, significantly outperforms both an extractive summarization method and a neural abstractive seq2seq model.

The specific contributions of our work are:

\begin{itemize}
    \item[-] We extend an encoder-decoder approach based on the Trans\-for\-mer-architecture, including an attention encoding that pays attention to salient comments.
    \item[-] We introduce a novel approach for text summarization. The summarization task's target includes detecting salient comments, extending prior single document approaches to a multi-document multi-author text summarization scenario.
    \item[-] We release a new dataset that includes news descriptions, reader comments, and likes, favoring reproducibility, and further extensions to our proposal. Data and codes are available here [anonymized for the double blind process].
\end{itemize}

The paper is organized as follows. In Section \ref{rel-work}, we discuss related work in text summarization with a specific focus on social context summarization. In Section \ref{proposal}, we introduce our model. The experimental methodology is discussed in Section \ref{methodology}. Experiments are reported in Section \ref{results}. Finally, we conclude in Section \ref{conc}, providing concluding remarks and discussing future work.

\section{Related Work}
\label{rel-work}
\subsection{Extractive and abstractive approaches}

Automatic text summarization techniques can be organized into two families of methods, extractive and abstractive methods \cite{shi2018neural}. On the one hand, extractive summarization methods choose text chunks (keywords, sentences, or paragraphs) that match a text body's semantic content \cite{JadhavR18}. Then, the summary corresponds to the extracted pieces of text. Iterative extractive models have extended these approaches by introducing the ability to generate multiple passes over the input to refine the text summary \cite{ChenGTSZY18}. The intuition behind these models is that multiple passes along the text may help the model polish the document representation. Selective reading mechanisms have been relevant in these approaches, improving the results obtained by single-pass extractive methods. Despite the advances obtained by these methods in text summarization, its main limitation remains being the lack of coherence that text chunks' concatenation can show.

On the other hand, abstractive summarization methods synthesize a text that matches a text body's semantic content \cite{BansalC18}. To abstract the text, the method uses a generative language model that synthesizes text that is consistent with the original text's content. Generally these models are based on the seq2seq architecture \cite{SutskeverVL14}. These methods may be more coherent but rely on more complex models. Accordingly, they must be carefully trained to produce consistent texts.


Many abstractive summarization methods are focused on summarizing single documents. For example, Subramanian et al.~\cite{subramanian2019extractive} present a novel approach for abstractive summarization of long scientific documents using a transformer-based architecture. The method outperforms seq2seq-based mechanisms according to ROU\-GE scores \cite{lin-2004-rouge}. The method considers two stages. First, using an extractive method, they select a few relevant sentences. Then, the method summarizes the chosen sentences using the transformer. Aside from ROU\-GE scores, the authors show that their generated summaries' abstractiveness is higher than those obtained using seq2seq mechanisms. The main limitation of this approach is its high computational cost. The authors report that the training process spent five days on a GPU-based machine. Bansal et al.~\cite{BansalC18} also combine an extractive method to select representative sentences from a document with an abstractive model. Both modules are combined hierarchically using a gradient method to bridge the non-differentiable computation between these two neural networks. Zhou et al.~\cite{ZhouYWZ17} use a seq2seq architecture to implement an abstractive summarization method. The authors introduce a selective sentence mechanism for sentence summarization in the encoder, including an attention layer in the model's decoder. By using these two variations on the seq2seq architecture, the authors show performance improvements in summarization tasks. Seq2seq architectures were extended with copy mechanisms, providing the model with the ability to select salient words or sentences from the input to be included in the model outcome. Zhou et al.~\cite{ZhouYWZ18} show that a seq2seq model endowed with copy mechanisms at sentence level improves the performance in summarization tasks.

Pretrained language models can be helpful for summarization tasks. Rosiello et al.~\cite{RossielloBS17} uses the centroid method for selecting sentences. The method's intuition is that a sentence is descriptive of a document if its text embedding is similar to the full text's centroid embedding. Experimental results show that this extractive method is very competitive. Its success is attributed to the use of pretrained language models based on word2vec \cite{MikolovSCCD13}. Liu et al.~\cite{liu2019text} combine a pretrained text encoder based on BERT with a transformer-decoder \cite{vaswani2017attention}. The authors report many fine-tuning strategies for both the encoder and decoder. Experimental results show that this proposal gets state-of-the-art scores in abstractive summarization.

\subsection{Summarization in Social Contexts}

Some summarization techniques applied to social media focus on solving single document problems. For instance, Lin et al.~\cite{LinSMS18} uses a convolutional neural network to include a text's global encoding, providing context to a seq2seq architecture. The authors show good results summarizing short posts collected from Sina Weibo. Wang et al.~\cite{WangYTZLD18} also uses the convolutional seq2seq architecture for text summarization, showing its success in many benchmark data.
Kim et al.~\cite{kim2018abstractive} propose an abstractive summarization method of short stories from Reddit. Their architecture, referred to as \emph{multi-level memory}, extracts representations of words, sentences, and paragraphs producing a hierarchical encoding of each story. The proposal gets good results, but the authors show that many of the summaries are biased towards each story's first paragraphs. This characteristic matches many of the gold standard summaries, which consider each document's first sentences a good outcome. This finding coincides with the results found by Kryscinski et al.~ \cite{kryscinski2019neural}, whose proving with human annotators how this bias can be exploited to get better metrics. These findings have pushed the focus of the problem towards the criticism of how the quality of an automatic summarization method is evaluated. For example, ROUGE \cite{lin-2004-rouge} has been criticized because it cannot distinguish between factual and subjective information, being also tackled by homographs \cite{shi2018neural}. 

Ying et al.~\cite{ying2015towards} summarize opinions from web forums. To fulfill this purpose, they introduce some features to filter out low-quality posts. These features consider lexical coverage, stylographic-text features, and subjectivity measured in terms of sentiment words. Using linear programming, they combine these features to implement an extractive summarization method. 

Gao et al.~\cite{gao2019comments} examine how to include reader comments to improve news articles descriptions. The authors study the distribution of information on comments incorporating these data into the summarization process. They introduce a seq2seq-based model named \emph{reader-aware summary generator} (RASG), which considers an attention layer that fed a semantic alignment score between the reader comments words and document words. The type of training used by RASG is based on adversarial learning. Experimental results show the feasibility of RASG improving the description of the news articles. 

Recently, Li et al.~\cite{li2020twitterabs} introduced an abstractive summarization model to describe social events on Twitter. The proposal uses a pretrained BERT encoder to code the tweets. Also, they fed an "Event Embedding" that captures the whole data stream. All this information is fed to a transformer-based decoder that generates the summary. The authors define a supervised task, grounded on human annotators that created a dataset of gold summaries for each event.

\vspace{2mm}

\noindent \textbf{Connection with the related work.} Many text summarization methods are based on the seq2seq architecture using single document single author approaches. When the social context is considered, some approaches pay attention to the description of an original subject, looking for alignments between the reader's comments and the original document. These models are based on single document approaches. Related work shows fewer efforts focused on identifying salient comments. Ying et al.~\cite{ying2015towards} use an extractive multi-document approach to identify these comments. Gao et al.~\cite{gao2019comments} and Li et al.~\cite{li2020twitterabs} propose abstractive approaches to address this problem but in single document contexts. Specifically, Gao et al.~\cite{gao2019comments} takes the reader's comments in a seq2seq architecture but discards the comments and the original document's misalignments. Li et al.~\cite{li2020twitterabs} is probably the closest approach to our proposal, as it uses the Transformer architecture to build event summaries on Twitter. However, it does not use users's likes to identify salient comments. Instead, the model is trained to generate human experts' target texts, focusing the model in generating a single document output that summarizes the whole event.

\section{Summarizing online news discussions}
\label{proposal}
\subsection{Motivation}

The summarization of online news discussions poses several challenges. Users' comments are generally in colloquial language. Accordingly, the extraction of relevant information from these messages requires an additional effort. Also, many times, users' comments can be uninformative. Many times, the news generate a massive flow of comments, many of them repetitive. A summarizer who considers the social context must grapple with these difficulties.

\begin{figure}[h!]
\vspace{-3mm}
    \centering
    \includegraphics[width=6cm]{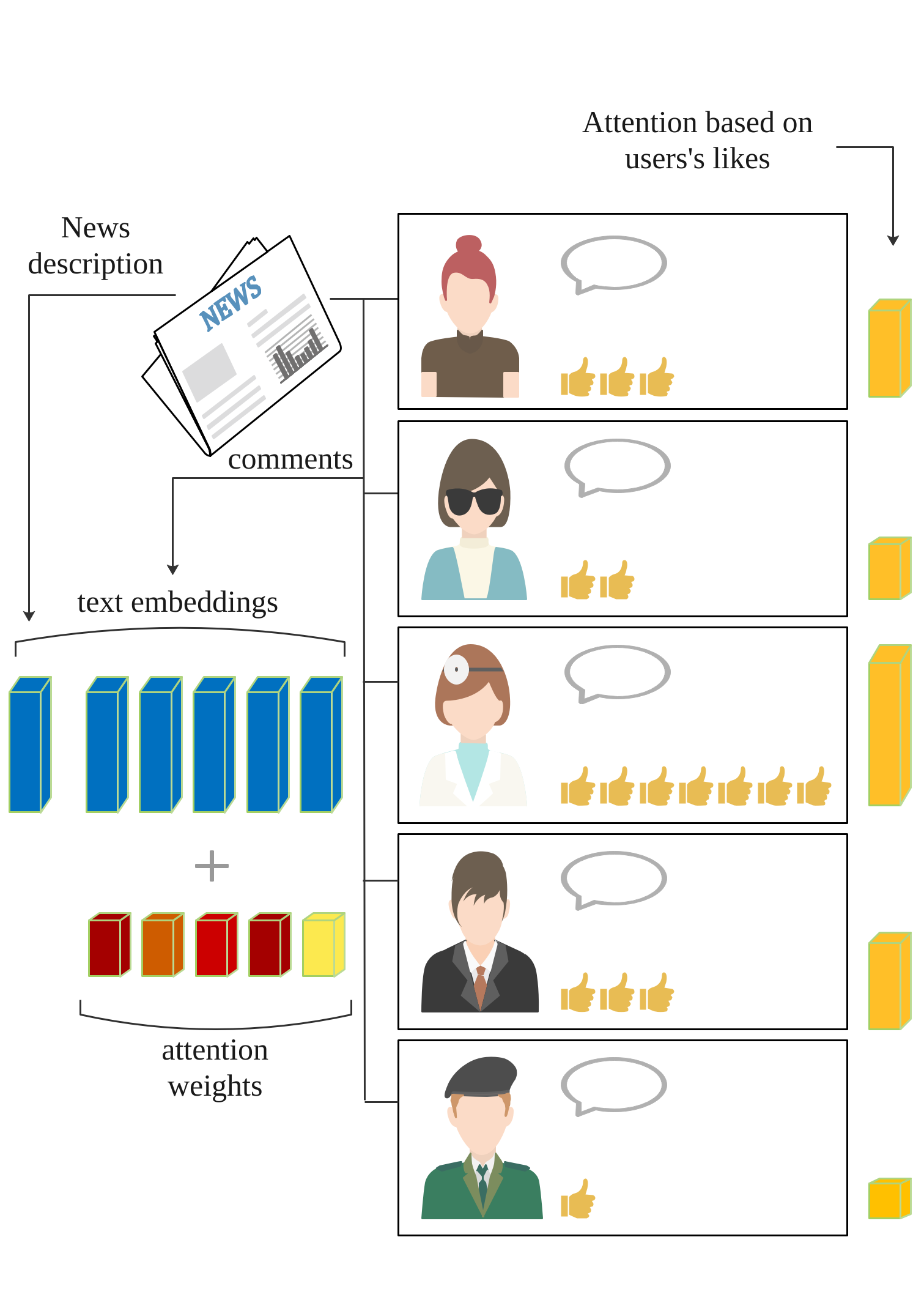}
    \vspace{-3mm}
    \caption{In our model, the attention weights are driven by the user's likes. The description of the news and the user's comments are embedded in the architecture using text embeddings. The architecture includes an attention encoding that prioritizes salient comments.}
    \label{fig:0}
\end{figure}

We claim that relevant information receives likes and that less informative comments attract less attention from users. This statement is correct if all messages have the same chance of being observed. In the case of digital news media, many comments are published on the same news site, and therefore, users can read these comments simultaneously as they review the descriptions of the news. With low editorial mediation, user comments can attract as much attention from the community as the original story in this type of site. In this scenario, we claim that likes reveal post informativeness. Accordingly, we propose to use the likes to drive the attention of an automatic summarization method to the most salient posts (see Figure \ref{fig:0}).

The key element of our proposal is the definition of a like-based attention encoding. We use a summarization architecture fed with news' descriptions (e.g., news' titles and subheads), and users' comments. Unlike related work, in which users' comments are selected in terms of their relevance to the news, we pay attention to the community's reaction. Figure \ref{fig:0} shows that comments with many likes will generate a greater attention weight than those comments with few likes. The attention based on users' likes will provide more information to the encoder, highlighting these comments during the encoding process. As these comments will be more relevant when coding the news and its comments, the decoder will reconstruct a social context summary giving more relevance to the comments with the most likes.

\subsection{Architecture}

Let $x_0$ be the news description provided by a digital media website, and let $x_1 \ldots x_n$ be the readers' comments. To feed these texts into a Transformer-based encoder, we build a text sequence. We organize a token sequence to keep the original structure of comments safe, providing a matrix of tokens with $n+1$ rows, one for each comment provided by the media. Using BERT's pretrained language model, we get a $(n+1~\times$~\texttt{tokens} $\times~768)$ 3D tensor, where each token is embedded with the $768$ dimensions provided by BERT. 

Let $\text{BERT}[x_0, \ldots, x_n]$ be the tensor constructed using BERT. 
$\text{BERT}[x_0, \ldots, x_n]$ is processed using the Transformer encoder. We use twelve transformer blocks to encode this input, obtaining a new vector encoding $\text{Enc}[x_0, \ldots, x_n]$. We combine $\text{Enc}[x_0, \ldots, x_n]$ with a social attention encoding before ingesting it into the decoder.

We define our social attention encoding using users' likes. The social attention encoding has $n+1$ components, one for each piece of text encoded in $\text{Enc}[x_0, \ldots, x_n]$. Each attention weight takes values in $[0, 1]$. For the news encoding $\text{Enc}[x_0]$, we define a maximum attention weight $\text{Att}[x_0] = 1$. For each comment $x_1, \ldots, x_n$, the attention is defined by:

\[
\text{Att}[x_i] = \sqrt{\frac{\text{likes}(x_i)}{\text{Max}_{likes}}}, \hspace{5mm} i \in \{1, \ldots , n \},
\]

\noindent where $\text{Max}_{likes}$ is the number of likes recorded by the most salient comment related to $x_0$. We compose $\text{Enc}[\ldots]$ and $\text{Att}[\ldots]$ using element-wise dot products. We use the Hadamard product for this purpose, getting a new encoding vector: 

\[
\text{Enc}_{\text{Att}}[x_i] = \text{Att}[x_i] \odot \text{Enc}[x_i], \hspace{5mm} i \in \{1, \ldots , n \}.
\]

The element-wise encoding strategy gives relevance to salient posts, diminishing the importance of comments with few likes. This encoding is then fed into the Transformer decoder to generate a new text. We use six decoder blocks for this task. The transformer uses a language model in its output to map the sequence generated by the decoder to the space of terms considered in the vocabulary. The decoder outcome is obtained by decoding the input provided by our social attention text encoding mechanism:

\[
y = \text{Dec}(\text{Enc}_{\text{Att}}[x_0,\ldots,x_n]).
\]

\begin{figure}[h!]
    \centering
    \includegraphics[width=7.5cm]{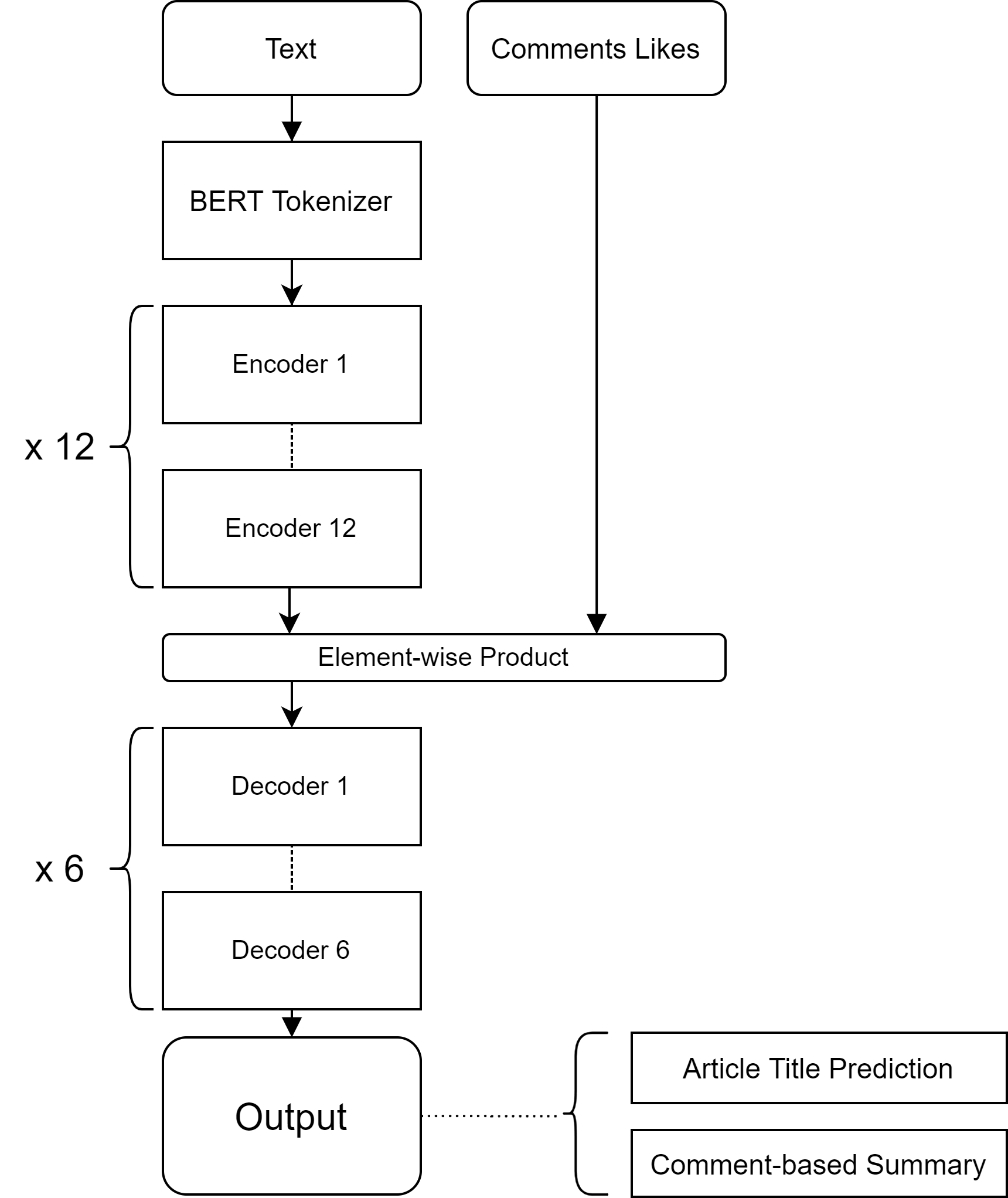}
    \caption{High-level model architecture. We use BERT to encode the text. Then, we process these data using the Transformer encoder. Attention weights are combined with text encodings, feeding them into the Transformer decoder.}
    \label{fig:hl_model}

\end{figure}

Figure \ref{fig:hl_model} shows the architecture proposed to solve the summarization tasks. To train this model, we will define a task that simultaneously generates the description of the news and the most salient comments.

\subsection{Model training}

To train the model, we define two tasks that are simultaneously addressed during the training stage. The first task is to generate comments randomly picked from the entry ($x_1$ to $x_n$) according to their attention weights. In this way, the model will learn to generate the most salient comments on each news. The second task considers the generation of the news title. In this way, the model will learn to predict the description of the news. We claim that by solving both tasks simultaneously, the decoder will relate comments and news descriptions.

We encode the training tasks as follows:

\[
y = \texttt{<News title> [UNUSED-10] <Chosen Comment(s)>}
\]

\noindent where $y$ is the target text, and it denotes the text decoded using our proposal. The number of comments included in the target is a parameter of the training stage. We evaluate different settings for this task, considering just one comment or multiple comments.

Note that as the training examples are generated using a random process, we reduce over-fitting chances. Also, we increase the number of available examples as we may run many iterations over the training partition producing different examples. More training iterations will produce a greater variety of training examples, improving the generalization capabilities of the model.

For training, we use the standard categorical cross-entropy loss function. We also include label smoothing minimizing the Kullback-Leibler divergence. This way of training the model is similar to the one used in the Open Neural Machine Translation project (Open NMT) \cite{klein-etal-2017-opennmt}. To handle the code generation at the model output, we use Beam Search \cite{wu2016googles} with five tokens. Each text sequence was decoded until the model outputs an \texttt{[EOS]} token. Repeated n-grams were blocked to prevent the generation of informal expressions \cite{liu2019text}.

\section{Design of the validation study}
\label{methodology}

\subsection{Data}

We work with real data retrieved from a digital news media [ano\-ny\-mized for the double-blind process] to study our model. The news media provides open data along with comments from its users. Also, users can vote for the comments of other users through the mechanism of likes. The data is openly published by the media using JSON format. The data was downloaded directly from the site. To comply with data privacy protocols, user identities were anonymized. The dataset is fully available in [ano\-ny\-mized for the double-blind process].

Users can create an account on the site or associate a Facebook or Twitter account within the website. All users with publishing rights do so with authentication. There are no anonymous comments. The website provides its users with mechanisms to comment on each news. Users can comment directly on a story. The site keeps a record of the comments related to each news item published by the digital media. The website provides a voting mechanism for comments, which considers likes. Users widely use this mechanism. Other interaction mechanisms are provided by the website, such as blocking users and reporting comments, which have less use. In addition to commenting on news, the users can keep a wall. In this place, opinions of general interest can be published. Users can follow other users to access their walls. However, the social graph associated with this interaction mechanism has low density, so it is of little interest to our study.

The entire dataset comprises 143340 news retrieved from April 1, 2016, to April 20, 2019. Of this set, 122778 news has comments. The news dataset contains different categories: National, International, Technology, Economy, Entertainment. We focused our work in the national news category, consisting of 41733 articles. This subset was partitioned into three folds: one for training with 35024 examples, one for validation consisting of 3354 news articles and 3355 articles for testing. The dataset partition was done randomly.

\subsection{Text Preprocessing}

The comments were cleaned using Python and \emph{regex} to remove any HTML tags and symbols. We remove people's mentions, URLs, laughter, and repeating punctuation. After this, comments with at least five words were included, reducing the number of news items on each partition of the dataset. We did not remove stopwords as they help with readability in the summary generation. The resulting dataset records 30024 news included in the training partition with 674937 comments. 2873 news were included in the validation partition with 62696 comments. Finally, 2874 news were included in the testing partition with 54080 comments.

Token sequences were generated using Standford CoreNLP4 \footnote{\ url{https://stanfordnlp.github.io/CoreNLP/}}. Finally, each word from the input was processed using the BERT subword tokenizer. This tokenizer splits words into subtokens from its pretrained vocabulary, mitigating the out of vocabulary problem. We worked with sequences of 512 subtokens, as is usual in BERT architectures \cite{devlin2018bert}.

\subsection{Experimental environment}

Every model tested was trained on [server specification anonymized for the double-blind process], 
using 16 cores of a Intel Xeon 6148 Skylake Processor, 32 GB of RAM and a GPU NVidia Tesla V100 SXM2 (16GB). 

We used five hours of computational time for each model checkpoint. Model checkpoints were saved and evaluated every 2000 steps, picking the top-scoring model for each of the training task variants shown in Table \ref{tab:tasks}.

\begin{table}[h]
\caption{Variants of our model evaluated in this study.}
    \centering
    \begin{tabular}{c|c|c|c}
ID &  Attention encoding & News title             & Chosen comment (s) \\ \hline
1  &  Disabled           & Enabled                & One comment \\
2  &  Disabled           & Disabled               & One comment \\
3  &  Enabled            & Enabled                & One comment \\
4  &  Enabled            & Disabled               & One comment \\
5  &  Disabled           & Enabled                & Three comments \\
6  &  Disabled           & Disabled               & Three comments \\
7  &  Enabled            & Enabled                & Three comments \\
8  &  Enabled            & Disabled               & Three comments \\ 
\end{tabular}
    
    \label{tab:tasks}
\end{table}

As Table \ref{tab:tasks} shows, we trained different variants of our model, enabling and disabling the attention encoding mechanism. We also study different task variants, considering tasks with and without title prediction. For the chosen comments, we study the prediction of one or three comments, all of them randomly picked from the input according to their attention weights.

\vspace{3mm}

\noindent \textbf{Tasks addressed in the testing instances.} To validate the models in the testing instances, we provide to each model the news title and the reader's comments. Users' likes were not provided to the models. In this way, the task that the model must address consists of, in addition to generating the summary from the text provided, identifying salient comments from the input. Variants with and without news titles were also considered, as indicated in Table 1.

\subsection{Validation metrics}

\begin{figure}[h!]
    \centering
    \includegraphics[width=6cm]{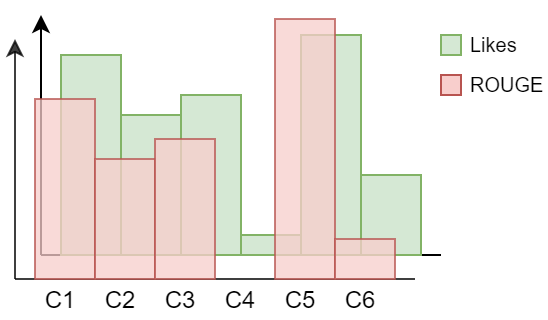}
    \caption{Distribution of ROUGE scores and likes obtained from a news item's comments and the text generated by our model. We expect that these distributions correlate, showing that the model can generate texts related to the comments that receive the most likes.}
    \label{fig:likesvrouge}
\end{figure}

To verify the relationship between model's outcomes and testing instances, we evaluate the model's performance computing the ROUGE score between the comments' summary and each comment on the news' thread. Accordingly, we obtain a ROUGE score distribution. We compare the obtained distribution with the distribution of likes received by news' comments. A well-performing model is expected to show similar distributions between likes and ROUGE scores, as it is shown in Figure \ref{fig:likesvrouge}.

We measure the match between both distributions computing the cross-entropy, which corresponds to a distributional measure named  XENT(Rouge). Since XENT(Rouge) is calculated using cross-entropy, better performance corresponds to values close to zero. 

In addition to using XENT(Rouge) to evaluate our models, we  further validate our results by computing a weighted average recall score of the comments, weighting them according to the amount of likes they received:
\[
Recall_w = \frac{\displaystyle \sum^n_{i=0} ROUGE_{recall}(x_i) \cdot likes(x_i)}{\sum likes(x_i)}
\]

Furthermore, we calculate the Recall ROUGE score between the news title and the generated summary. By doing so, we obtain performance measures for how relevant is the summary when compared to the discussion topic of the thread.

As we test variants of our model disabling the attention encoding or the title prediction task, we compare the performance achieved with and without the indicated variant using a paired T-test. With these tests, we check if the difference between both variants is statistically significant.


\section{Experiments}
\label{results}

\subsection{Experimental results}

We show in Tables \ref{tab:single_results} and \ref{tab:tri_results} the results obtained for predicting one comment or three comments, respectively. We use two state-of-the-art methods to compare the results obtained by our model with its closest competitors. We selected two methods, one extractive and one abstractive. The extractive method corresponds to the centroid-based model proposed by Rossiello et al.~ \cite{RossielloBS17}. This method was selected as it allows us to evaluate our model's capacity against one based on pretrained language models with an extractive approach. In this way, we can evaluate our model's contribution to an extractive method built on the same language model. We used BERT to implement the centroid-based model. The other state-of-the-art method selected for the validation section was the abstractive technique proposed by Shi et al.~\cite{shi2018neural}. In this way, we can evaluate our model's contribution against a state-of-the-art method based on the seq2seq architecture.

\begin{table}[htbp]
\caption{Average scores for XENT(Rouge), Weighted Recall and mean ROUGE scores over the news article's title prediction, for the single comment prediction task variants.}
\centering
\begin{tabular}{@{}lccc@{}}
\toprule
\multicolumn{4}{c}{\textbf{One comment}}         \\ \midrule
Model                               & XENT(Rouge)    & $Recall_w$     & ROUGE          \\ \midrule
Extractive \cite{RossielloBS17}    & 0.462          & 0.515          & 0.438          \\
Abstractive \cite{shi2018neural}    & 0.386          & 0.594          & 0.518          \\ \hline
\textbf{Model-(1)}                           & \textbf{0.358} & \textbf{0.613} & \textbf{0.555} \\
Model-(2)                           & 0.494          & 0.510          & 0.458          \\
Model-(3)                           & 0.372          & 0.582          & 0.548          \\
Model-(4)                           & 0.395          & 0.596          & 0.530          \\ \bottomrule
\end{tabular}
\label{tab:single_results}
\end{table}

\begin{table}[htbp]
\caption{Average scores for XENT(Rouge), Weighted Recall and mean ROUGE scores over the news article's title, for the prediction of three comments.}
\centering
\begin{tabular}{@{}lccc@{}}
\toprule
\multicolumn{4}{c}{\textbf{Three comments}}                                         \\ \midrule
Model                            & XENT(Rouge)    & $Recall_w$     & ROUGE          \\ \midrule
Extractive \cite{RossielloBS17} & 0.178          & 0.781          & 0.692          \\
Abstractive \cite{shi2018neural} & 0.187          & 0.770          & 0.711          \\ \hline
Model-(5)                        & 0.176          & 0.798          & 0.734          \\
Model-(6)                        & 0.243          & 0.727          & 0.668          \\
\textbf{Model-(7)}                        & \textbf{0.153} & \textbf{0.838} & \textbf{0.783} \\
Model-(8)                        & 0.175          & 0.827          & 0.779          \\
 \bottomrule
\end{tabular}
\label{tab:tri_results}
\end{table}

Table \ref{tab:single_results} shows that when generating only one comment and the news title, the best model variant is Model-(1). Model-(1) disables the attention encoding of likes and enables the title prediction task together with the prediction of a salient comment. This variant of the model achieves the best XENT(Rouge) (lowest) and the highest ROUGE score and weighted recall. The other three variants of the model are not effective in this task. Table \ref{tab:single_results} also shows that our model outperforms both state-of-the-art models. The differences between Model-(1) and the rest of the methods are statistically significant (p $<$ 0.05, paired t-test). Table \ref{tab:single_results} also shows that when activating attention encoding, the performance of Model-(3) declines concerning Model-(1) in both metrics. This result means that the attention encoding variant does not contribute relevant information in the news title + single comment generation task. This may be because the model prioritizes the alignment between the title text and the most salient comment, which could be enough to solve this task.

Table \ref{tab:tri_results} shows that when generating three comments and the news title, the best model variant is Model-(7). Model-(7) uses attention encoding, making the model obtain the best XENT(Rouge) and the highest ROUGE score. The other three variants of the model are less effective in this task. Table \ref{tab:tri_results} also shows that our model outperforms both state-of-the-art models. The differences between Model-(7) and the rest of the methods are statistically significant (p $<$ 0.05, paired t-test). The results also show that by activating the attention encoding, it is possible to improve both tasks, significantly reducing the XENT(Rouge) and the ROUGE score concerning Model-(5), trained on the same task but with attention encoding disabled. When disabling the news title prediction task and the attention encoding, the worst result is obtained in XENT(Rouge), which shows that generating many comments requires the attention encoding.

\subsection{Discussion of results}


We characterize the best and worst results of our models. To do this, we partitioned the testing instances into quartiles, with Q1 and Q4 being the quartiles that have the instances with the best and worst XENT(Rouge) scores achieved by our models. To characterize these instances, we measured different characteristics, such as the length of the news thread, the number of comments, and the length of the comment(s) to be generated. We also measure the lexical diversity of the news thread and the salient comment(s). For each quartile, we computed averaged features across testing instances. Finally, we also measure the standard deviation of the distribution of likes. These results are shown for our best models, Model-(1) and Model-(7), in Tables \ref{tab:single_metrics} and \ref{tab:tri_metrics}, respectively.

\begin{table}[h]
\caption{Characterization of the best and worst results achieved by Model-(1). Testing instances were sorted according to their XENT(Rouge) scores. LD indicates lexical diversity and STD standard deviation.}
\centering
\begin{tabular}{lrr}
\hline
\multicolumn{3}{c}{\textbf{One comment}}            \\ \hline
Feature                       & Q1          & Q4    \\ \hline
Length of the thread     & 386         & 322   \\
Length of the salient comment & 84          & 21    \\
\# of comments                & 21          & 17    \\
LD of the thread         & 53\%        & 58\%  \\
LD of the salient comment     & 72\%        & 91\%  \\
STD of likes                  & 0.217       & 0.187 \\ \hline
\end{tabular}
\label{tab:single_metrics}
\end{table}

\begin{table}[h]
\caption{Characterization of the best and worst results achieved by Model-(7). Testing instances were sorted according to their XENT(Rouge) scores. LD indicates lexical diversity and STD standard deviation.}
\centering
\begin{tabular}{lrr}
\hline
\multicolumn{3}{c}{\textbf{Three comments}} \\ \hline
Feature                      & Q1          & Q4         \\ \hline
Length of the thread               & 404         & 267        \\
Length of the salient comments            & 165         & 68         \\
\# of comments              & 21          & 14         \\
LD of the thread                   & 52\%        & 60\%       \\
LD of the salient comments                & 63\%        & 76\%       \\
STD of likes       & 0.218       & 0.179      \\ \hline
\end{tabular}
\label{tab:tri_metrics}
\end{table}

Table \ref{tab:single_metrics} shows that shorter threads are more difficult for Model-(1). This characteristic is also expressed in the number of comments. On average, the most difficult instances have 15 comments. The more difficult examples also have a distribution of likes with a lower standard deviation. These examples also show a low lexical use (higher LD), showing that lexical use in these types of examples is also poorer. These trends are also observed when evaluating Model-(7), as it is shown in Table \ref{tab:tri_metrics}. For this model, the most difficult instances (Q4) also correspond to shorter news threads, with fewer comments and poorer lexical use. This evaluation also shows that the most difficult instances have a lower standard deviation than those of Q1.

\subsection{Illustrative examples}

In Figure \ref{fig:examples}, we show some illustrative examples that help in understanding how our model works. We picked two examples from the testing partition, at random, showing the news title, a brief explanation of the context, some illustrative comments, and the social context summarization provided by our model. We highlight in magenta the comments to which our model paid more attention to generate the summary. Some less relevant comments are shown in yellow.

\begin{figure}[ht!]
\noindent
\begin{minipage}[t]{.48\textwidth}
\footnotesize
\raggedright
\textbf{News title}: \textit{The government supports his Minister and emphasizes that his comments to the Museum do not represent him nowadays} \\
\vspace{2mm}
\textbf{Context}: \textit{A minister argues against the museum of human rights} \\
\vspace{2mm}
\textbf{Comments}:\\
- \colorbox{magenta}{User 1}: \textit{And what is the problem that he thinks like that, just like millions of people?} $\{$likes: 74$\}$ \\
- \colorbox{yellow}{User 2}: \textit{Hey Lone ranger, of what millions are you talking about, when all of our country abhorred HH.RR. violations} $\{$likes: 1$\}$ \\
- \colorbox{magenta}{User 3}: \textit{It would be valuable for visitors to comprehend, for example, that between many of the factors that contributed to the destruction of democracy, prevailing violence had a significant role.} $\{$likes: 10$\}$ \\
- \colorbox{magenta}{User 4}: \textit{In June of 2012, the Director of libraries manifested her discomfort with the museum because on her opinion it offers "an incomplete look of the facts"} $\{$likes: 6$\}$ \\
- \colorbox{magenta}{User 5}: \textit{<Anonymized> is not the first a collaborator of the government that thinks critically of the museum.} $\{$likes: 10$\}$ \\
- \colorbox{yellow}{User 6}: \textit{It is the museum that is wrong and only shows one of the sides of the coin.} $\{$likes: 2$\}$ \\
\vspace{2mm}
\textbf{Social context summarization}: \textit{It would be a great contribution that the museum explained the situation before the coup. In June of 2012, he manifested his discomfort with the museum"} \\
\centering{(a)}
\end{minipage}%
\hfill
\begin{minipage}[t]{.48\textwidth}
\footnotesize
\raggedright
\textbf{News title}: \textit{Politician comments about not presenting in the constitutional Court: "It does not offer any guarantee of legitimacy"} \\
\vspace{2mm}
\textbf{Context}: \textit{A politician comments on his decision not to attend court} \\
\vspace{2mm}
\textbf{Comments}:\\
\vspace{3mm}
- \colorbox{yellow}{User 1}: \textit{I believed him more when he was an "objective" journalist.} $\{$likes: 10$\}$ \\
- \colorbox{yellow}{User 2}: \textit{The constitutional Court, inspiration for ours, is completely politicized, it has been one of the reasons for the government to want independence from <anonymized>.} $\{$likes: 10$\}$ \\
- \colorbox{yellow}{User 3}: \textit{In our country is no different} $\{$likes: 18$\}$ \\
- \colorbox{magenta}{User 4}: \textit{It's a political body not legal.} $\{$likes: 40$\}$ \\
- \colorbox{magenta}{User 5}: \textit{The constitutional reform of 2005 says that the constitutional court will be composed by 3 members chosen by the supreme court in secret ...} $\{$likes: 29$\}$ \\
- \colorbox{magenta}{User 6}: \textit{I think it would be positive if the process for naming its members is modified by establishing a mechanism that does not involve so much interference from political power...} $\{$likes: 1$\}$ \\
- \colorbox{magenta}{User 7}: \textit{He disqualifies one of the members of the constitutional Court because offer any guarantee of legitimacy or seriousness, ....} $\{$likes: 1$\}$ \\
\vspace{2mm}
\textbf{Social context summarization}: \textit{The constitutional reform of 2005 says that the constitutional court will be composed by 3 members... I believe to change the process for naming its members ... He disqualifies one of the members of the constitutional Court...} \\
\centering{(b)}
\end{minipage}

\caption{Examples of news, comments, and the summarization provided by our method.}
\label{fig:examples}
\end{figure}

Example a) is used in single comment tasks and b) in multi comment tasks. Our model achieves good results in a) and b). Example a) shows controversial news that produces opposed views among users. The community of this media gives more likes to the comments that support the government and less to those against the government. The most relevant comment corresponds to user 1, who receives 74 likes. The comments of users 3, 4, and 5 also receive likes. The comments of users 2 and 6 are much less relevant to this media community and receive very few likes. When generating the social context summarization, the model extracts text from the most salient comments, generating a long text that summarizes a good part of the users' opinions who support the government. The model discarded comments in yellow. This example shows that attention encoding allows the model to focus on users' most relevant comments, discarding those that record few likes. It can be seen in this example that the comments have lexical richness. The model pays attention to this factor and generates a very informative summary of the discussion. Another successful example of our model is the one shown in Figure \ref{fig:examples} b). The generated summary corresponds to a multi comment task. We can also observe that the discussion's lexical richness is important, which is used by the model to generate an informative summary of the discussion. However, unlike the previous example, the model does not pay attention to some comments with likes, such as comments in yellow. This is because user 1's comment is about the politician, not about the subject of the news. Something similar happens with the comments of users 2 and 3, who compare the context of the news with the situation observed in another country. These three comments have in common that they do not refer directly to the news subject but to tangential issues. The comments of users 4, 5, 6, and 7 refer to the news subject and therefore are more relevant when generating the summary. In this example, the attention encoding was not so relevant for the model, but rather the comments' alignment with the news title. This finding relieves the fact that our model obtains its best result when solving both tasks simultaneously: news title generation and multi comment generation. Solving both tasks simultaneously helps the model to find comments that are aligned with the subject of the news. 

\subsection{Main findings and limitations of our model}

Experimental results show that our model behaves well in cases where the news discussion has lexical richness. We have also observed that, in general, shorter threads have longer salient comments and longer threads have shorter salient comments. This finding suggests that there is an invariant in terms of information distribution in social discussions. Our data show that there are discussions with very little lexical richness and others with high lexical richness. The proposed model takes advantage of the lexical richness to generate an informative summary. If the discussion is poor, so is the summary. 
Likes encoding allows the model to prioritize the comments with the most likes. This mechanism is successful if the discussion is lexically rich. If the discussion is lexically poor, the model is unable to generate an adequate summary. This is because the model has limitations in relating one discussion to other discussions. The model operates by constructing a summary by paying close attention to the news's comments but little attention to the comments associated with similar news. The model also shows some risks. In cases where the news produces controversial points of view, the attention encoding mechanism prioritizes a view, the one with the most likes, discarding comments from minorities. This bias towards the most voted comments can introduce an over-representation of the dominant points of view, minimizing the influence of minority groups that contribute with a greater diversity of opinions to the discussion. The generation of summaries of social contexts that do not amplify these biases is a relevant challenge for this model type.

\section{Conclusions}
\label{conc}
Unsupervised text summarization allows us to extend the range of settings where standard text summarization cannot work due to the lack of human-made golden summaries. The use of user comments to generate these standards bring challenges such as slang abuse and lexical typos. We have developed a transformer-based model that allows us to summarize news articles' comments while considering the social context that these interactions live in. We compare our model with state-of-the-art methods. Our experimental results show that our model performs better than its competitors as it can take advantage of the popularity of comments. In particular, our model takes advantage of discussions when they are lexically rich.

%
Future work aims to tackle coverage of different opinions that do not necessarily represent the majority. Processing the data to detect opinion clusters could extend our training task to promote diversity in comments, giving a more balanced view of the users' opinions on social platforms. Additionally, this work reveals the need to develop validation metrics to address ROUGE's limitations. Finally, coding interactions between users, using for example graph neural networks, can incorporate valuable information into the representation used by the model. We will explore the use of these types of representations to evaluate their impact on social context summarization.

\section*{Acknowledgements}
Mr. Tampe acknowledges funding from the Emerging Leaders in the Americas Program (ELAP) and Dalhousie University. Mr. Mendoza acknowledges funding from the Millennium Institute for Foundational Research on Data. Mr. Mendoza was also funded by ANID PIA/APOYO AFB180002 and ANID FONDECYT 1200211.

\bibliographystyle{unsrt}  
\bibliography{references}  

\end{document}